\documentclass[runningheads]{llncs}

\usepackage{graphicx}
\usepackage{amsmath,amsfonts,amssymb}
\usepackage{siunitx}
\usepackage{subfig}
\usepackage{bbm}
\usepackage{hyperref}

\begin{document}
\title{Robust Conformal Volume Estimation in 3D Medical Images}
\author{Benjamin Lambert \inst{1, 2} \and
Florence Forbes \inst{3} \and Senan Doyle \inst{2} \and Michel Dojat \inst{1}}
\authorrunning{B. Lambert et al.}

\institute{Univ. Grenoble Alpes, Inserm, U1216, Grenoble Institut Neurosciences, 38000, FR \and
Pixyl, Research and Development Laboratory, 38000 Grenoble, FR \and
Univ. Grenoble Alpes, Inria, CNRS, Grenoble INP, LJK, 38000 Grenoble, FR}


\maketitle              
\begin{abstract}
Volumetry is one of the principal downstream applications of 3D medical image segmentation, for example, to detect abnormal tissue growth or for surgery planning. Conformal Prediction is a promising framework for uncertainty quantification, providing calibrated predictive intervals associated with automatic volume measurements. However, this methodology is based on the hypothesis that calibration and test samples are exchangeable, an assumption that is in practice often violated in medical image applications. A weighted formulation of Conformal Prediction can be framed to mitigate this issue, but its empirical investigation in the medical domain is still lacking. A potential reason is that it relies on the estimation of the density ratio between the calibration and test distributions, which is likely to be intractable in scenarios involving high-dimensional data. To circumvent this, we propose an efficient approach for density ratio estimation relying on the compressed latent representations generated by the segmentation model. Our experiments demonstrate the efficiency of our approach to reduce the coverage error in the presence of covariate shifts, in both synthetic and real-world settings. Our implementation is available at \url{https://github.com/benolmbrt/wcp_miccai}.

\keywords{Uncertainty  \and Predictive Interval \and Covariate Shift}
\end{abstract}

\section{Introduction}
\label{sec:intro}

An important downstream application of medical image segmentation is the extraction of volume measurements for lesions or organs. Lesion volumetry plays a pivotal role in various medical scenarios, including predicting the outcome after stroke \cite{ghoneem2022association}, grading brain tumors \cite{baris2016role}, or monitoring the progression of Multiple Sclerosis \cite{mattiesing2022spatio}. Brain volumetry can also be useful to monitor atrophy \cite{contador2021longitudinal}, and analyzing the volume of organs is useful for aging studies \cite{wasserthal2023totalsegmentator}. However, automated segmentations can be error-prone, which inevitably leads to imprecise volumetric measurements. A potential solution would be to associate predictive intervals (PIs) with the estimations to take into account this uncertainty. 

Conformal prediction (CP) \cite{lei2015conformal,papadopoulos2002inductive} is an uncertainty paradigm allowing to associate PIs with regressed scores (here, volumes). The most popular variant of CP, Split CP \cite{papadopoulos2002inductive}, relies on a set-aside calibration dataset (generally a subset of the training dataset) that is used to calibrate the intervals so that they match the target coverage level on fresh test data. However, it is based on the exchangeability hypothesis, following which calibration and test data are drawn independently from the same distribution. In general, this is not the case for medical image processing applications, where domain shifts are extremely common, due to variations in the data acquisition protocol or the presence of pathologies unseen during training \cite{xue2023cross}. When calibration and test data points are not exchangeable, the accuracy of the conformal procedure collapses drastically \cite{barber2023conformal,tibshirani2019conformal}, which hinders the relevancy of conformalized PIs in medical applications.

As a potential solution, Weighted Conformal Prediction (WCP) has been proposed to account for shifts between calibration and test distributions \cite{barber2023conformal,tibshirani2019conformal}. It is based on the reweighting of calibration samples according to the estimated density ratio $dP_{\text{test}} / dP_{\text{train}}$. As a result, calibration samples close to the test samples are attributed with higher importance in the conformal procedure. A flourishing literature can be found for density ratio estimation, with popular approaches including the training of a classifier to distinguish between training and test distributions \cite{bickel2007discriminative,angelopoulos2022conformal}, moment \cite{gretton2009covariate} or ratio matching \cite{kanamori2009least}. More recently, Deep Learning (DL) approaches are also investigated to estimate density ratios \cite{ding2020subsampling,nam2015direct}. However, we note that applications of WCP to medical image segmentation are still lacking, which may be due to the difficulty of estimating the density ratio for high-dimensional imaging data. 

In this work, we propose to investigate the use of WCP to tackle covariate shifts in medical image segmentation tasks, with the ultimate goal of computing calibrated PIs for lesion volumes. As a contribution, we propose an efficient way of computing the density ratio in high-dimensional medical images, by relying on latent representations generated by the segmentation model. 

\section{Conformal Prediction for volumetry in medical images}

\subsection{Problem definition}
We consider a 3D segmentation problem with $N$ classes where our objective is to estimate the true volumes $Y\in \mathbb{R}^{N-1}$ of each foreground class based on the predicted segmentation. Within this framework, for an estimation $X$ of the volume, we define a predictive interval $\Gamma_{\alpha}(X)$ as a range of values that are constructed to contain the true volume $Y$ with a user-defined degree of confidence $1-\alpha$ (e.g $90\%$ or $95\%$). More formally, given a set of estimated volumes $X_{1} \ldots X_n$ and their corresponding ground truth volumes $Y_1 \ldots Y_n$, $\Gamma_{\alpha}(\cdot)$ should be learned such that it satisfies \cite{angelopoulos2021gentle}:

\begin{equation}\label{eq_ch4_marginal}
    1-\alpha \leq P(Y_{\text{test}}\in\Gamma_{\alpha}(X_{\text{test}})) \leq 1-\alpha + \frac{1}{n+1}
\end{equation}

\subsection{Predictive Interval computation using a multi-head segmentation architecture}

\begin{figure}
    \centering
    \includegraphics[width=0.8\textwidth]{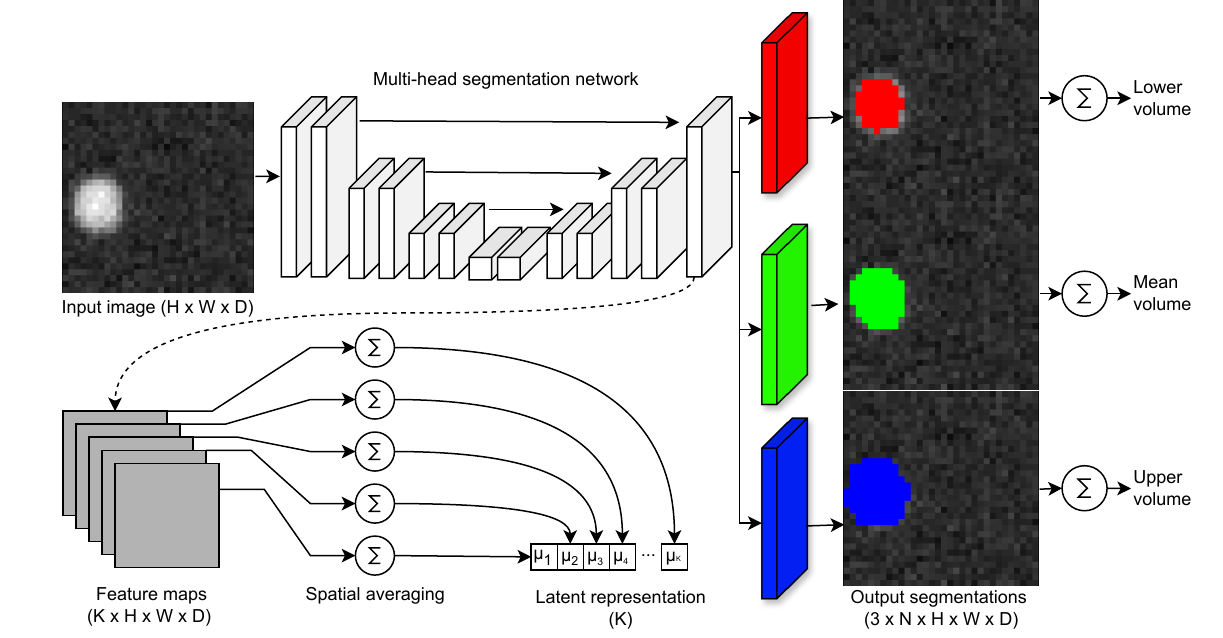}
    \caption{Illustration of the proposed framework. A multi-head segmentation model predicts three distinct masks for each label: a restrictive mask associated with the lower bound volume (red), a permissive mask associated with the upper bound volume (blue), and a balanced mask for the average volume (green). For Weighted Conformal Prediction, a compressed latent representation is extracted from the penultimate convolution filter. }
    \label{fig:framework}
\end{figure}

In practice, a PI associated with a volume $X_i$ is composed of a lower bound $l_i$, and an upper bound $u_i$. For a practical estimation of these three quantities ($X_i$, $l_i$ and $u_i$), \cite{lambert2023triadnet} proposed to train a multi-head segmentation network that predicts 3 output masks: a restrictive one (low recall, high precision) to estimate the lower bound, a permissive one (high recall, low precision) to estimate the upper bound, and a balanced one for the estimation of the mean (Figure \ref{fig:framework}). The key element is to perform training using the Tversky loss $T_{\alpha, \beta}$ \cite{salehi2017tversky} allowing to control the penalties applied to false positives (FP) and negatives (FN) contained in each mask through the loss parameters $\alpha$ and $\beta$, respectively. Writing $p_{\textit{lower}}$, $p_{\textit{mean}}$ and $p_{\textit{upper}}$ the outputs of each head and $y$ the ground-truth segmentation, the loss is defined:

\begin{equation}\label{eq_loss}
    \mathcal{L} = T_{1-\gamma, \gamma}(p_{\textit{lower}, y}) + T_{0.5, 0.5}(p_{\textit{mean}, y}) + T_{\gamma, 1-\gamma}(p_{\textit{upper}, y})
\end{equation}

where $\gamma$ is a hyperparameter set to $0.2$ controlling the penalties applied to FP and FN during the training of the lower and upper bound heads.

To ensure that the computed PIs will achieve the user-defined level of coverage on test data, the conformal calibration of intervals can be performed \cite{angelopoulos2021gentle}. It operates by first defining a score function $s_i = \max(l_i - Y_i, Y_i - u_i)$. This score is a way to estimate the accuracy of the interval $[l_i, u_i]$ for the true quantity $Y_i$, with larger scores indicating larger discrepancy. The scores are computed on a set-aside calibration dataset comprising $n$ pairs of images and associated ground truths. It allows to compute the $\frac{\lceil(n+1)(1-\alpha)\rceil}{n}$-th quantile of the empiral scores: $\hat{q} = \text{Quantile}(s_1, s_2, ..., s_n ; \frac{\lceil(n+1)(1-\alpha)\rceil}{n})$. In practice, $\hat{q}$ acts as a corrective factor applied to the PIs so that they encompass the desired fraction of the true volumes on the calibration dataset. At test time, the calibrated PI is computed as follows:

\begin{equation}
            \Gamma_{\alpha}(X_i) = [l_i - \hat{q}, u_i + \hat{q}]
\end{equation}

As a result, the intervals expand as  $\hat{q}$ increases. Supposing the test samples are exchangeable with the calibration samples, the marginal coverage property (Equation \ref{eq_ch4_marginal}) is guaranteed.  

\subsection{Weighted Conformal Prediction to tackle covariate shift}

WCP has been proposed to take into account the non-exchangeability of calibration and test data \cite{angelopoulos2021gentle,barber2023conformal,tibshirani2019conformal}. The core concept of WCP is to reweight the calibration dataset to more accurately match the test one. This is achieved by estimating the density ratio $w=dP_{\text{test}}/dP_{\text{train}}$ for each calibration and test sample. In practice, writing $X_1, ..., X_n$ the $n$ calibration samples and $x$ the fresh test point, importance weights are computed as:

\begin{align}
    p_i^w(x) &= \frac{w(X_i)}{\sum_{i=1}^N w(X_j) + w(x)} 
\end{align}\label{eq_piw}

Essentially, the weight is large when the calibration sample $X_i$ is likely under the test distribution. Then, the corrective value $\hat{q}$ can be reframed as the $1-\alpha$ quantile of the reweighted distribution \cite{angelopoulos2021gentle}: 

\begin{equation}
    \hat{q}(x) = \inf \biggl\{ s_j: \sum_{j=1}^n p_i^w(x) \mathbbm{1} \{s_i \leq s_j \} \geq 1 - \alpha \biggr\}
\end{equation}\label{eq_qhat_weighted}

Note that when all weights are equal to $\frac{1}{n+1}$, the standard CP procedure is recovered. A convenient way to estimate this ratio is to use an auxiliary classifier that only requires that unlabeled samples from the test distribution are available during the calibration step \cite{tibshirani2019conformal}. The idea is to train a probabilistic classification model to classify samples between the training and test distributions. That is, writing $X_1, ..., X_n$ and $X_{n+1}, ..., X_{n+m}$ the training and test data points, one can form a classification dataset composed of the pairs $\{X_i, C_i\}$ where $C_i = 0$ for $i=1, ..., n$ and $C_i = 1$ for $i=n+1, ..., n+m$. Writing $\widehat{p}(x) = \mathcal{P}(C=1 \vert X =x)$ the probability predicted by a classifier model trained on the $\{X_i, C_i\}$ dataset that the input sample $x$ belongs to the test distribution, the weight function can be expressed as \cite{sugiyama2010density}:

\begin{equation}\label{eq:density_ratio_classif}
    \widehat{w}(x) = \frac{\widehat{p}(x)}{1-\widehat{p}(x)}
\end{equation}

However, this approach has several limitations. First, it requires access to a sufficient amount of calibration and test samples to allow for a supervised classification strategy. Second, training the classifier is cumbersome when dealing with high-dimensional medical images. In this setting, the dedicated classification approach would be the training of a deep learning Convolution Neural Network (CNN), requiring numerous examples of both classes (calibration and test). Moreover, the training of the auxiliary classifier should be performed during the CP procedure to allow for weight estimation. Incorporating the training of a CNN in the CP procedure is thus highly inefficient. As a conclusion, this classification task is computationally too costly when dealing with 3D medical images. Building on these limitations, we next investigate a more efficient approach making use of the latent representations extracted by the deployed segmentation model.

\subsection{Efficient density ratio estimation using latent representations}

As training the auxiliary classifier directly from the input images is too costly, more efficient approaches have to be investigated. One idea would be to use a compressed representation of the input image that still preserves important structural information. A lead in this direction is the use of low-dimensional latent representations generated by the segmentation model during the inference process, which has been proven to be a highly efficient summary allowing the detection of out-of-distribution images \cite{calli2022frodo,gonzalez2022distance,anthony2023use,woodland2023dimensionality}. Therefore, using compressed latent representations in place of the high-dimensional 3D images seems promising as our end goal is to address covariate shifts. To test this framework, we collect the activations of the penultimate convolution layer. The feature maps have a shape of $K \times H \times W \times D$, where $H$, $W$, and $D$ are the spatial dimensions of the 3D image and $K$ the number of kernels in the layer. This feature map is reduced to a compressed vector $z$ of dimension $K$ by performing an averaging over the spatial dimensions (see Figure \ref{fig:framework}). This approach allows training the auxiliary classifier on compressed representations of the input MRIs, which can be performed efficiently during the WCP procedure. 


\section{Experiments}
\subsection{Synthetic dataset with controlled covariate shift}
\begin{figure}
    \centering
    \includegraphics[width=0.8\textwidth]{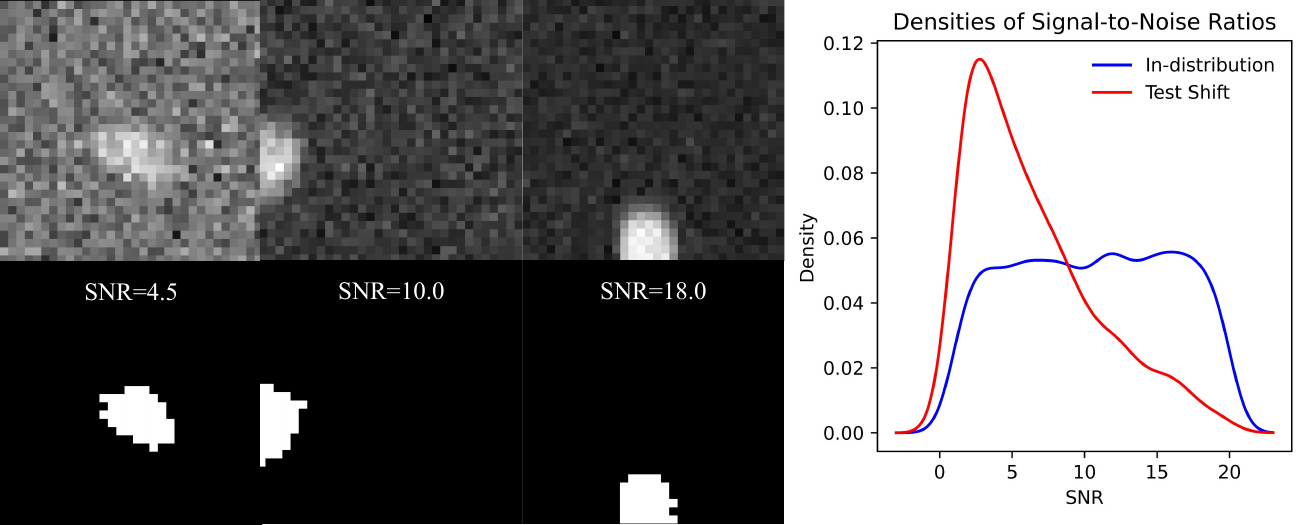}
    \caption{Left: Examples of synthetic images with varying Signal-to-Noise ratios (SNRs) and associated ground truths. Right: Distribution of SNRs in the in-distribution and shifted synthetic datasets. }
    \label{fig:synthetic}
\end{figure}

To prove the relevancy of the proposed approach, we first rely on a synthetic setting allowing us to control covariate shift precisely. The task that we propose here is the segmentation of spheres inside cubic volumes of shape $32\times32\times32$, with the end goal of computing a PI for the volume of each sphere. The covariate of interest here is the signal-to-noise ratio (SNR) between the background of the image and the foreground spheres. A total of $4000$ synthetic images are generated. We then split this dataset into an in-distribution (ID) split (3000 images) containing images with high SNRs, and a shifted test dataset (1000 images) containing images with lower SNRs. The ID dataset is further split into training, calibration, and ID test parts, with 1000 images each. Several examples of synthetic images with varying SNRs are presented in Figure \ref{fig:synthetic}, along with the densities of SNR in the ID and shifted datasets.

\subsection{Real-world covariate shift in brain tumor segmentation tasks}

To test the framework on real-world medical image data, we address multi-class tumor segmentation in brain MRI. Our dataset consists of glioblastoma and meningioma subjects gathered from the open-source BraTS 2023 datasets \cite{labella2023asnr,menze2014multimodal}. Each subject has four MRI sequences: T1-weighted, T2-weighted, FLAIR, and T1 with contrast enhancement. Ground truth masks include necrosis, edematous, and gadolinium-enhancing tumor classes. For covariate shift analysis, we divide subjects into an ID dataset (320 glioblastoma subjects, 748 meningioma subjects, $30\%$-$70\%$ repartition) and a shifted test dataset (873 glioblastoma subjects, 196 meningioma subjects, $82\%-18\%$ repartition). The covariate shift thus corresponds to the difference in frequencies of each subtype of tumor in the ID and shifted datasets. The ID dataset is further divided into training (568), calibration (250), and ID test (250) subsets.

\subsection{Experimental Protocol and Metrics}
We use MONAI's \cite{the_monai_consortium_2020_4323059} Dynamic U-Net \cite{futrega2021optimized} as segmentation backbone, modified to have three output heads. The penultimate convolution layer contains $64$ kernels, meaning that the extracted latent representations will also have a dimension of $64$. The models are trained using Equation \ref{eq_loss} and the ADAM optimizer \cite{kingma2014adam} with a learning rate of $2\times 10^{-4}$. After training, PIs are calibrated on the calibration dataset, with a target coverage of $95\%$ for the synthetic task, and $90\%$ for brain tumors as the segmentation is more challenging. Three variants of CP are further compared:

\begin{itemize}
    \item \textbf{Standard} CP corresponds to the setting where calibration samples are associated with identical weights, thus not taking into account potential covariate shifts.
    \item Weighted CP using Oracle covariates (\textbf{W-Oracle}) uses the ground truth covariates to train the auxiliary classifier: SNR for the synthetic images, tumor subtype for brain tumors (0 for glioblastoma, 1 for meningioma). 
    \item Weighted CP using latent representations  (\textbf{W-Latent}) leverages the compressed latent representations to train the auxiliary classifier. 
\end{itemize}

For W-Oracle and W-Latent, we use a Logistic Regression (LR) model as the auxiliary classifier, trained in a 20-fold cross-validation setting. The probabilities predicted by LR are clipped in the range $[0.01, 0.99]$ to avoid infinite weights (see Equation \ref{eq:density_ratio_classif}). To estimate the performance of the CP procedures, we monitor the empirical coverage on the test datasets (ID and shifted) as well as the average interval width. We also report the segmentation performance using Dice scores, and the accuracy of the auxiliary LR classifier for W-Oracle and W-Latent. The experiments are reproduced for $R=250$ trials by shuffling the ID calibration and test datasets. The shifted test dataset is kept identical in each trial. 

\section{Results and Discussion}
Tables \ref{tab:effective_size_accuracy_CP_syn} and \ref{tab:effective_size_accuracy_CP_glio} present the performance of each CP variant on the synthetic and brain tumor datasets, respectively. In the absence of covariate shifts (ID datasets), W-Oracle and W-Latent closely mimic Standard CP, achieving target coverages with great accuracy ($95\%$ for synthetic data, $90\%$ for brain tumors). However, in Shift datasets, Standard CP exhibits miscoverage, with empirical coverages lower than the target level, revealing its inability to handle non-exchangeable data points. W-Oracle and W-Latent alleviate this issue, with W-Oracle recovering the exact target coverages on the synthetic task and the necrosis and edematous brain tumor classes. W-Latent also reduces the coverage gap, although it doesn't exactly recover the target coverages. It can be noticed that this increase robustness is linked with an increase in the average interval width to achieve the target coverage on shifted test data.

\begin{figure}%
    \centering
    \subfloat[\centering Synthetic dataset]{{\includegraphics[width=0.45\textwidth]{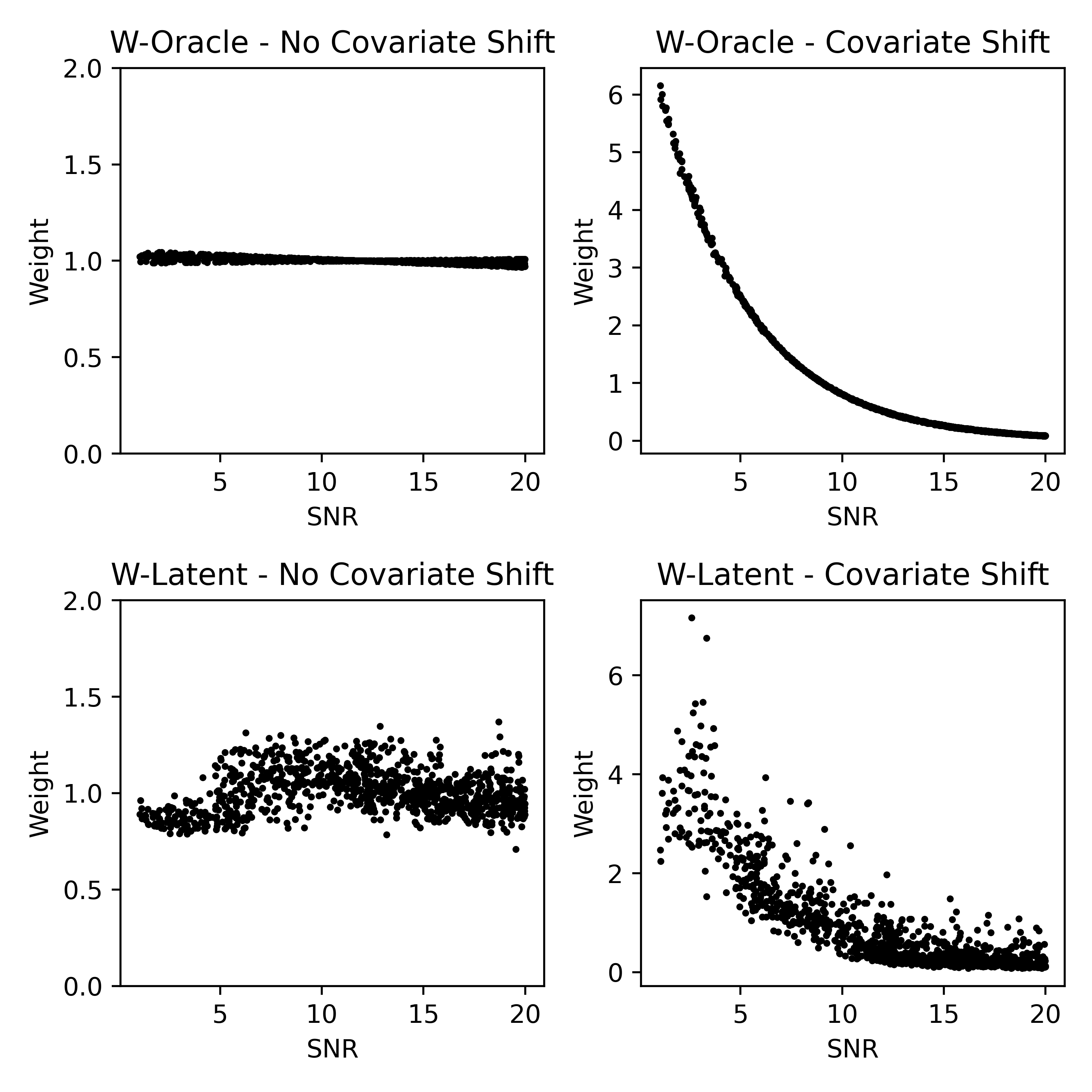} }}%
    \qquad
    \subfloat[\centering Brain tumor dataset ]{{\includegraphics[width=0.45\textwidth]{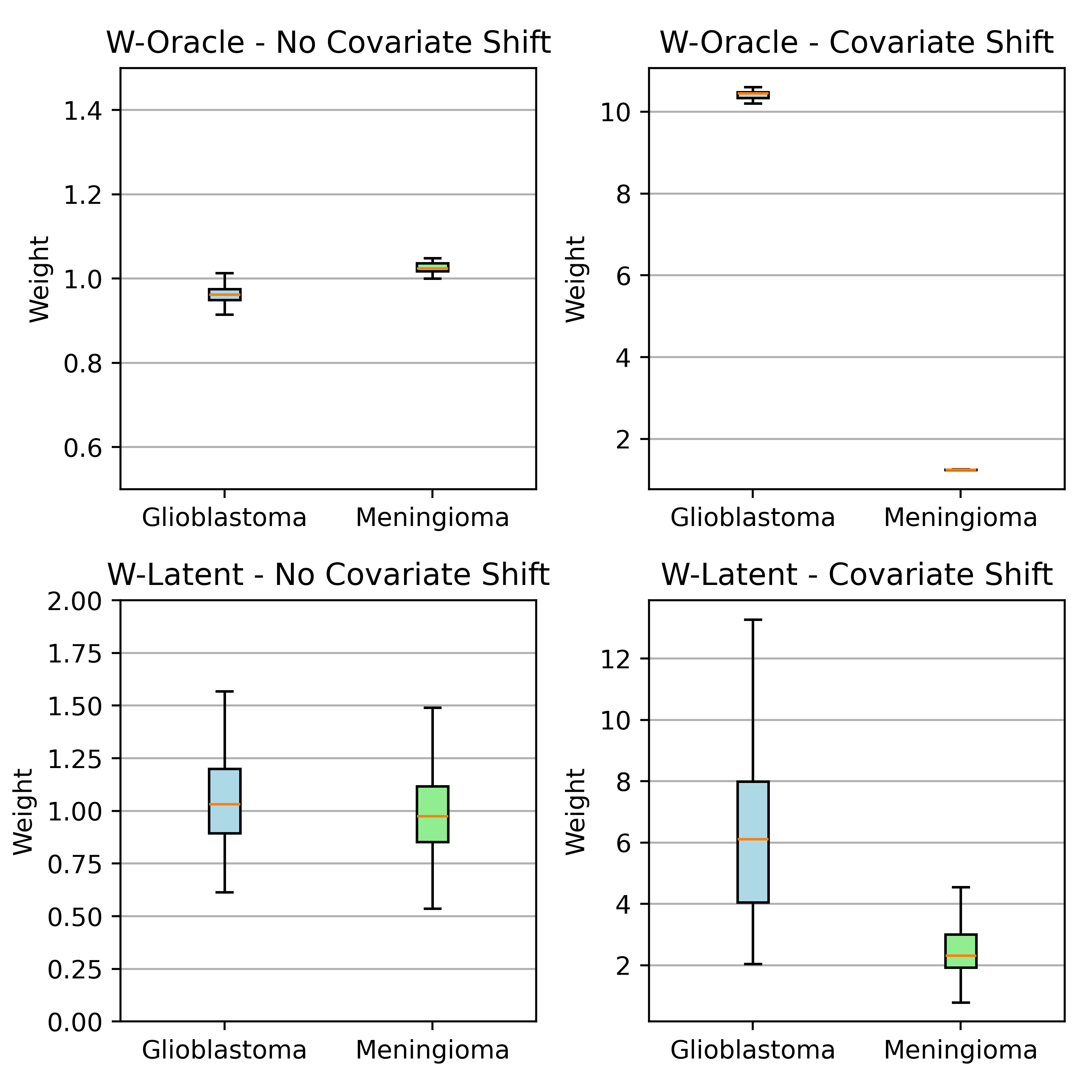} }}%
    \caption{Weights of calibration samples estimated by W-Oracle and W-Latent, with and without covariate shift, according to the value of the covariate.}%
    \label{fig:weights}%
\end{figure}

A deeper dive into the functioning of WCP is presented in Figure \ref{fig:weights}. It presents the calibration weights' behavior with and without covariate shifts. When there are no shifts, all weights are close to the unit, mimicking the standard CP procedure. When a covariate shift is observed, higher weights are assigned to calibration samples resembling test samples. For the synthetic task, higher weights are attributed to calibration weights with low SNRs, which are similar to the shifted test samples. For tumor segmentation, higher weights are attributed to glioblastoma subjects, which indeed represent the majority of the shifted test subjects. W-Oracle and W-Latent provide similar trends, although W-Latent is more noisy than the Oracle version. 

In conclusion, our WCP framework is effective in tackling covariate shifts in medical image analysis, by addressing covariate shifts either directly or through latent representations, ensuring the robustness of predictive intervals. However, one limitation of the presented WCP framework is that it can only account for moderate covariate shifts. Otherwise, if the covariate shift is too important between calibration and test samples, the weights will likely diverge (see Equation \ref{eq:density_ratio_classif} when $\hat{p}(x)$ converges to 1) which would undermine the accuracy of the WCP procedure. 

\begin{table}[]
    \centering
    \scriptsize
    \caption{Comparison of standard and weighted Conformal Prediction on the synthetic task, for a target coverage of $95\%$. The mean and standard deviation over 250 trials are presented.}
    \begin{tabular}{c|c|c|c|c|c}
        \textbf{Setting} & \textbf{CP version} & \textbf{Accuracy}  & \textbf{Coverage} ($\%$) & \textbf{Width} (\SI{}{\cubic\milli\meter}) & \textbf{Dice} \\ \hline 
         Calib ID & Standard &  - & $95.11 \pm 0.93$ & $86.18 \pm 3.00$ & \\ 
         vs. & W-Oracle & $0.50 \pm 0.02$ & $95.01 \pm 1.01$ & \textbf{$86.06 \pm 2.95$} & $0.92 \pm 0.07$ \\  
         Test ID & W-Latent  & $0.50 \pm 0.01$ & $95.03 \pm 0.93$ & $86.45 \pm 3.79$ &\\  \hline 
         
         Calib ID & Standard &  - & $87.47 \pm 1.08$ & \textbf{$94.76 \pm 2.77$} &\\ 
         vs. & W-Oracle & $0.73 \pm 0.01$  & \textbf{$95.22 \pm 1.57$} & $153.28 \pm 23.52$ & $0.89 \pm 0.11$ \\ 
         Test Shift & W-Latent & $0.72 \pm 0.01$ & $93.39 \pm 0.90$ & $128.89 \pm 11.34$&  \\
         
    \end{tabular}
\label{tab:effective_size_accuracy_CP_syn}
\end{table}

\begin{table}[]
    \centering
    \caption{Comparison of standard and weighted Conformal Prediction on multi-class tumor volume estimation for a target coverage of $90\%$. The mean and standard deviation over 250 trials are presented.}
    \scriptsize
    \begin{tabular}{c|c|c|c|c|c|c}
          \textbf{Class} & \textbf{Setting} & \textbf{CP version}  & \textbf{Accuracy} & \textbf{Coverage} ($\%$) & \textbf{Width} (\SI{}{\milli\litre}) & Dice \\ \hline 
          &  Calib ID & Standard & - & $90.40 \pm 2.62$ & $4.1 \pm 0.7$ & \\ 
          & vs. & W-Oracle & $0.50 \pm 0.04$ & $90.09 \pm 2.67$ & $3.9 \pm 0.7$ &  $0.63 \pm 0.33$ \\
         Necrosis & Test ID & W-Latent & $0.50 \pm 0.03$ & $90.20 \pm 2.61$ & $4.0 \pm 0.7$  & \\ \cline{2-7} 

          & Calib ID & Standard & - & $80.19 \pm 2.28$ & $6.2 \pm 1.0$ & \\ 
          & vs. & W-Oracle  & $0.70 \pm 0.01$ & $89.53 \pm 2.45$ & $13.0 \pm 2.6$  & $0.71 \pm 0.28$  \\
         & Test Shift & W-Latent & $0.81 \pm 0.00$ & $88.88 \pm 2.88$ & $12.4 \pm 3.0$ & \\  \hline 

         & Calib ID & Standard & - & $90.48 \pm 2.77$ & $18.9 \pm  1.8$ & \\ 
        & vs. & W-Oracle & $0.50 \pm 0.04$ & $90.19 \pm  2.85$ & $18.6 \pm 1.9$ & $0.81 \pm 0.22$  \\
         Edematous & Test ID & W-Latent & $0.50 \pm 0.03$ & $90.29 \pm 2.81$ & $18.7 \pm 2.0$ & \\ \cline{2-7} 

          & Calib ID & Standard & - & $80.56 \pm 2.35$ & $26.2 \pm 2.2$ & \\ 
          & vs. & W-Oracle & $0.70 \pm 0.01$& $89.58 \pm 2.44$ & $39.7 \pm 5.4$ & $0.80 \pm 0.20$  \\
         & Test Shift & W-Latent & $0.81 \pm 0.00$ & $85.52 \pm 3.91$ & $32.9 \pm 6.0$  & \\  \hline 

         & Calib ID & Standard & - & $90.54 \pm 2.57$ & $5.0 \pm 0.4$ & \\ 
          & vs. & W-Oracle & $0.50 \pm 0.04$ &  $90.29 \pm 2.64$ & $4.9 \pm 0.4$ &  $0.88 \pm 0.20$ \\
         GD-enhancing & Test Shift & W-Latent & $0.50 \pm 0.03$ & $90.36 \pm 2.55$ & $4.9 \pm 0.4$ & \\ \cline{2-7} 

          & Calib ID & Standard & - & $81.29 \pm 1.76$ & $7.0 \pm 0.4$ & \\ 
          & vs. & W-Oracle  & $0.70 \pm 0.01$ & $87.25 \pm 3.39$ & $9.0 \pm 1.5$ & $0.85 \pm 0.18$  \\
         & Test Shift & W-Latent & $0.81 \pm 0.00$ & $86.19 \pm 3.90$ & $8.6 \pm 1.6$  & \\

    \end{tabular}
    \label{tab:effective_size_accuracy_CP_glio}
\end{table}

\bibliographystyle{splncs04}
\bibliography{biblio}

\end{document}